\documentclass[10pt,twocolumn,letterpaper]{article}

\usepackage{cvpr}
\usepackage{times}
\usepackage{epsfig}
\usepackage{graphicx}
\usepackage{amsmath}
\usepackage{amssymb}

\usepackage[pagebackref=true,breaklinks=true,letterpaper=true,colorlinks,bookmarks=false]{hyperref}

\cvprfinalcopy

\ifcvprfinal\fi

\newcommand{\R}{\mathbb{R}}
\newcommand{\N}{\mathbb{N}}

\begin{document}

\title{Chargrid-OCR: End-to-end Trainable Optical Character Recognition for Printed Documents using Instance Segmentation}

\author{Christian Reisswig \thanks{Equal contribution}\\
\and
Anoop R Katti \footnotemark[1]\\
\and
Marco Spinaci \footnotemark[1]\\
\and
Johannes H\"ohne \footnotemark[1]\\
{\small SAP, Berlin Germany } \\
{\scriptsize \{christian.reisswig $\vert$ anoop.raveendra.katti  $\vert$ marco.spinaci $\vert$ johannes.hoehne \}@sap.com}
}

\maketitle

\begin{abstract}
We present an end-to-end trainable approach for Optical Character Recognition (OCR) on printed documents. 
Specifically, we propose a model that predicts a) a two-dimensional character grid (\emph{chargrid}) representation of a document image as a semantic segmentation task and b) character boxes for delineating character instances as an object detection task. 
For training the model, we build two large-scale datasets without resorting to any manual annotation - synthetic documents with clean labels and real documents with noisy labels.
We demonstrate experimentally that our method, trained on the combination of these datasets, (i) outperforms previous state-of-the-art approaches in accuracy (ii) is easily parallelizable on GPU and is, therefore, significantly faster and (iii) is easy to train and adapt to a new domain.
\end{abstract}

\section{Introduction}

Optical Character Recognition (OCR) on documents is a problem for which numerous open-source (e.g.~\cite{tesseractgit}) as well as proprietary~\cite{abbyy} solutions exist. However, this does not imply that the task itself is solved. The biggest challenges are: dense text organized in highly varying document layouts on the one hand and extremely high accuracy requirements on the other hand~\cite{breuel2017robust}. Fig.~\ref{fig:ex_full_out} visualizes an example document. In order to meet these challenges, the current state-of-the-art document-level OCR solutions consist of complex pipelines, where each step is either a hand-optimized heuristic or requires intermediate data and annotations to train the models.

Deep neural networks have been proven very successful in a number of computer vision tasks. Of particular interest for our approach to OCR are the task of spotting object instances using object detection~\cite{SSD} and recognizing (i.e. classifying) different regions in an image using semantic image segmentation~\cite{badrinarayanan2017segnet}. These two tasks, when performed together are also referred to as instance segmentation~\cite{he2017mask}. 

Inspired by these developments, in this work, we treat OCR on documents as detecting and recognizing character instances on a page. More concretely, we predict a character grid or \textit{chargrid} representation \cite{chargrid} of the input document, realized as a semantic segmentation task. Further, to delineate different character instances, we predict character boxes realized as an object detection task. By means of this re-formulation of the OCR task, we introduce a new end-to-end trainable OCR solution for printed documents that is based on fully convolutional neural networks. We refer to our method as \emph{Chargrid-OCR}.

In contrast to instance segmentation, a document image may contain several thousand characters on a page. We, therefore, consider our approach to tackle \emph{ultra-dense} instance-segmentation. The mainstream box filtering algorithm, Non-Maximum Suppression (NMS)~\cite{SSD, FasterRCNN, lin2017focal, he2017mask}, being a quadratic-time algorithm, does not scale for our ultra-dense scenario. Hence, we introduce a preliminary step, called \emph{Graphcore}, that runs in linear time and significantly speeds up box filtering. In addition to instance segmentation, as is common in OCR, characters should to be aggregated into words. Keeping in mind the scale and arbitrary rotation of words, we introduce a linear time clustering-based step to construct words from characters.

While there exists a number of large-scale datasets for training standard instance segmentation models, to the best of our knowledge, datasets for document OCR at a similar scale do not exist. We build two large-scale datasets with no manual annotation - one out of synthetic documents (with clean labels) built from Wikipedia content and another out of real financial documents, with noisy labels coming from the state-of-the-art open source OCR solution, Tesseract v4~\cite{tesseractgit}. We show that the two datasets complement each other and bring improvements when combined. More importantly, training on the two datasets, we demonstrate that we are able to out-perform the state-of-the-art thereby entirely avoiding any manual annotation. 
In addition to outperforming the state-of-the-art in accuracy, our method is easily parallelizable on GPU and therefore can be significantly faster. Further, our method is easier to train, both from scratch and for fine-tuning on an unseen domain.

Our main contributions include
\emph{(1)} end-to-end trainable model for document OCR
\emph{(2)} efficient post-processing for ultra-dense text extraction
\emph{(3)} building two large-scale document OCR datasets that complement each other without resorting to any manual annotation
\emph{(4)} pushing the accuracy of document OCR while being significantly faster and easier to train as well as adapt

\begin{figure*}[ht]
    \centering
    \includegraphics[width=\linewidth]{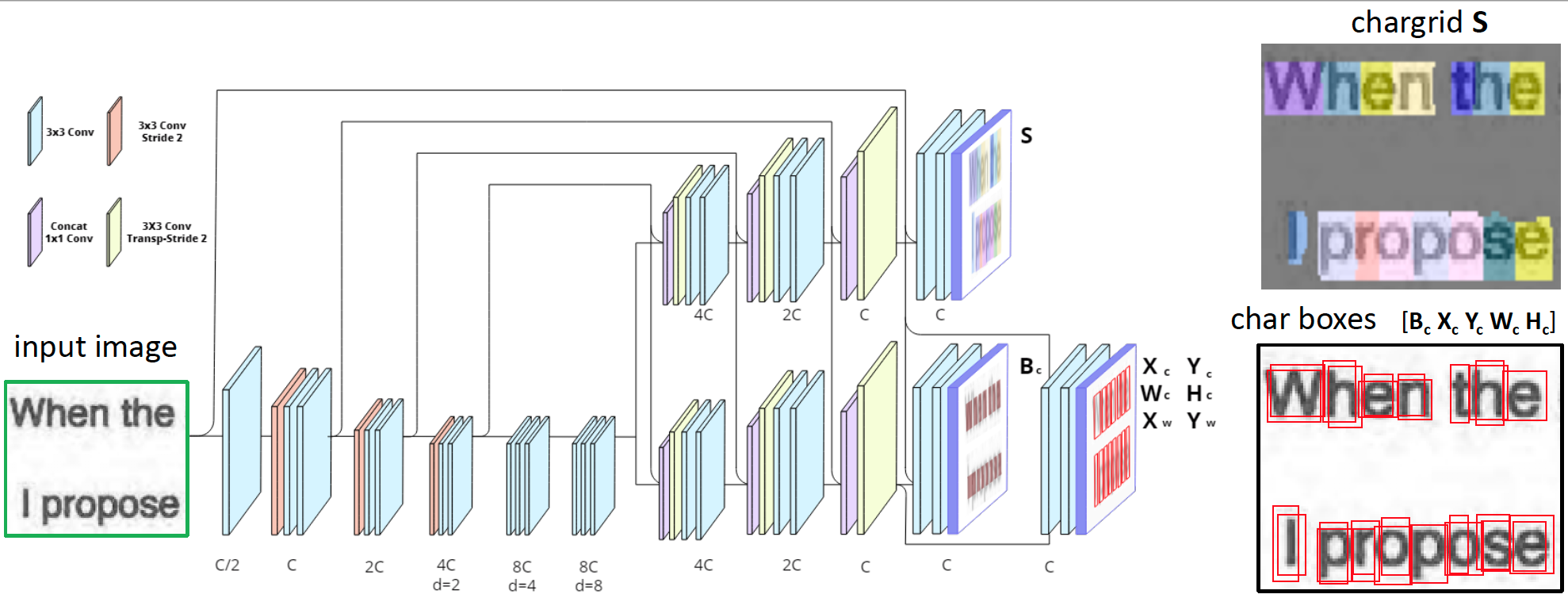}
    \caption{Chargrid-OCR architecture with an example input and predictions. For visualization purposes, only a small crop of a page is depicted as input / output. Parameters $nC$ and $d$ denote the number of channels and strides per convolution filter. $C$ is referred to as base channels.}
    \label{fig:architecture}
\end{figure*}

\section{Related Work}
State-of-the-art document OCR~\cite{tesseractgit, abbyy} are made of pipelines of several processing steps. The most popular is perhaps Tesseract~\cite{smith2007tesseract, smith2009hybrid, smith2009adapting}~\footnote{https://github.com/tesseract-ocr/tesseract}. Typical steps include binarization, skew correction, layout analysis / line segmentation and text recognition. The main reason for this pipeline complexity is to deal with dense text organized in a highly varying layout combined with high accuracy requirements. 
There have been several works that have tried simplifying the pipeline as well as turning each step into a trainable module~\cite{breuel2013high, breuel2017high, breuel2017robust, breuel2019tutorial}.
Furthermore, Tesseract itself has been under active development by incorporating LSTM-based text recognition into its latest major release in October 2018.

Such pipelines have the advantage, that each step is a well defined problem which can be optimized individually. However, a sequential OCR framework can suffer from the fact that errors in early processing steps can impact all following modules \cite{yousefi2015binarization}. Moreover, it may be cumbersome to train such a sequence of modules on new data such as new languages, document layouts, fonts, noise sources and so on. Furthermore, it requires a lot of effort to maintain such systems.

Interestingly, the very first application of Convolutional Neural Networks (CNNs) was character recognition, albeit in a much more restricted scenario~\cite{lecun1998gradient}. We build on a long line of work in deep object detection and deep semantic segmentation~\cite{krizhevsky2012imagenet, 43442, badrinarayanan2017segnet, SSD, lin2017focal} 
and propose an end-to-end trainable document OCR system.

Similar trends are being observed in the related task of Scene Text Detection and Recognition (STD and STR respectively), where pipeline-based methods~\cite{zhou2017east, panhe17singleshot, Jianqi17RRPN, shi2016end, cheng2018aon, lee2016recursive} have been recently replaced with end-to-end trainable models~\cite{xing2019charnet, qin2019towards}. A parallel work called CharNet~\cite{xing2019charnet} is perhaps the closest to ours. CharNet also predicts character boxes and character classes mainly to cope with arbitrarily shaped text commonly observed in natural images. This further validates our approach for document OCR. The main difference with our work lies in datasets for training the model and post-processing for extracting the text. While they train on synthetic data and then iteratively transfer to \emph{manually annotated} real data, we build two imperfect but complementing datasets with \emph{no manual annotations}. Further, our efficient post-processing steps are explicitly designed for handling extraction of large amounts of text that is observed in printed documents.

\section{Chargrid-OCR: End-to-end trainable OCR}
\label{sec:method}

We formulate document OCR as an instance segmentation task~\cite{he2017mask} on characters. Specifically, we perform semantic segmentation of the document image with characters as labels. The resulting segmentation mask is also referred to as the chargrid representation of the document~\cite{chargrid}.
Chargrid, by itself, does not allow one to delineate character instances. Therefore, we further use class agnostic object detection to predict bounding boxes for each character. 
The characters are further aggregated into words. This is achieved by predicting word centers at each pixel and clustering the characters based on the predicted word centers.
Due to directly predicting character segmentation mask, character boxes and word centers from the document image, our method is (i) lexicon free, i.e. only character based (ii) end-to-end trainable and finally, (iii) easily parallelizable on GPU.

\subsection{Model}
The input to our model is an image, $I \in \R^{H\times W}$, with text (e.g.~a printed document), where $H$ and $W$ are the height and the width of the image.
The output character segmentation mask, chargrid, $S \in \N^{H\times W}$, classifies each pixel in the input image into characters (Fig.~\ref{fig:architecture} top right). The characters are encoded by positive integers, e.g. ``A'' $\mapsto 1$, ``B'' $\mapsto 2$, etc., with $0$ being reserved for background. For English, our model uses 89 such symbols (26 lower-case, 26 upper-case, 10 digits, 26 between punctuation and special characters, plus a special ``unknown'' token.)
The character boxes are represented with box detection mask and box centers, widths and heights. This is similar to existing object detection methods~\cite{SSD, FasterRCNN}. The box detection mask is a binary mask, $B_c \in \R^{H\times W}$, denoting the presence of a character box on each pixel. 
The box centers, $(X_c, Y_c) \in (\R^{H\times W}, \R^{H\times W})$, predict the offset from each pixel to the center of the predicted character box that lies on that pixel. 
The widths and heights, $(W_c, H_c) \in (\R^{H\times W}, \R^{H\times W})$, predict the log-width and log-height of the predicted character box.

We further predict word centers, $(X_w, Y_w) \in (\R^{H\times W}, \R^{H\times W})$. In order to handle the large variance of the word widths and heights, they are encoded as $(X_w(i, j), Y_w(i, j)) = \Big(\text{sign}(\Delta x_w) \cdot \log\big(|\Delta x_w|+1\big), \text{sign}(\Delta y_w) \cdot \log\big(|\Delta y_w|+1\big) \Big)$, where $(\Delta x_w, \Delta y_w)$ denote the offsets from pixel $(i, j)$ to the center of the predicted word box that lies on that pixel.

We have a few thousand characters and a few hundred words on a page. In order to perform character detection at this scale, we use a single-stage approach, similar to e.g.~\cite{SSD}.
The architecture of our model is based on a fully-convolutional encoder-decoder structure, with one encoder and two decoders - one for semantic segmentation, another for boxes, branching out of the common encoder. Fig.~\ref{fig:architecture} illustrates the architecture with an example input and its corresponding outputs.
The encoder consists of several convolutional blocks, each comprised of a sequence of convolutional layers. We use stride-2 convolutions in the first three convolutional encoder blocks to decrease resolution by up to a factor of 8. In the deeper convolutional encoder blocks, we apply dilated convolutions (up to $d=8$). 
In the decoder blocks, we use transposed convolutions to increase resolution again. Each decoder block is connected via a skip connection to its corresponding encoder block of the same resolution. Thus our network bears resemblance to the original U-Net architecture \cite{UNET}. 
$C$, in Fig.~\ref{fig:architecture}, denotes the configurable width of the network, i.e. the number of channels in each layer. It is referred to as the \emph{base-channel} of the network.
Batch Normalization~\cite{ioffe2015batch} and ReLu activations~\cite{krizhevsky2012imagenet} are applied after each intermediate convolution. Spatial dropout~\cite{tompson2015efficient} is applied just before any skip connections, in both encoders and decoders.
The weights are initialized following \cite{DelvingDeepIntoRectifiers}.

We remark at this point that there is no need for the output resolution to match the input one, or even to have the same shape. Indeed, we found it beneficial to skip the last one or two upsampling steps, especially along the $y$ direction. The skip connections going from a higher to a lower resolution are also replaced by convolutions with adequate strides (and kernel size equal to the stride).

The model is trained using categorical cross-entropy~\cite{badrinarayanan2017segnet} for the segmentation outputs ($S, B_c$) and using Huber loss~\cite{FasterRCNN} for the regression output ($X_c, Y_c, W_c, H_c, X_w, Y_w$). We use Stochastic Gradient Descent with momentum for training, with momentum value 0.9, learning rate of $10^{-2}$ for the first 360K steps and $10^{-3}$ for the next 360K steps and a batch size of 2.

\subsection{Post-processing}
The outputs of the network are (i) character segmentation mask (i.e. chargrid) (ii) character box detection mask (iii) character box centers, width and heights and (iv) word box centers. Similar to standard object detection, character box detection mask often makes redundant predictions around the same character (as shown in Fig.~\ref{fig:architecture} bottom right). Such predictions need to be filtered to arrive at a single box per character. Further, characters need to be aggregated into words. 
However, there can be on the order of $10^3$ characters on a page and correspondingly, on the order of $10^5$ (redundantly) predicted character boxes. This is ultra-dense when compared to what is typically encountered in object detection in natural images. 
Below, we describe our solution for \emph{efficiently} extracting character boxes from the neural network outputs and aggregating them into words.

\begin{figure}
    \centering
    \includegraphics[width=0.44\textwidth, trim={0cm 11.5cm 0cm 7.7cm},clip]{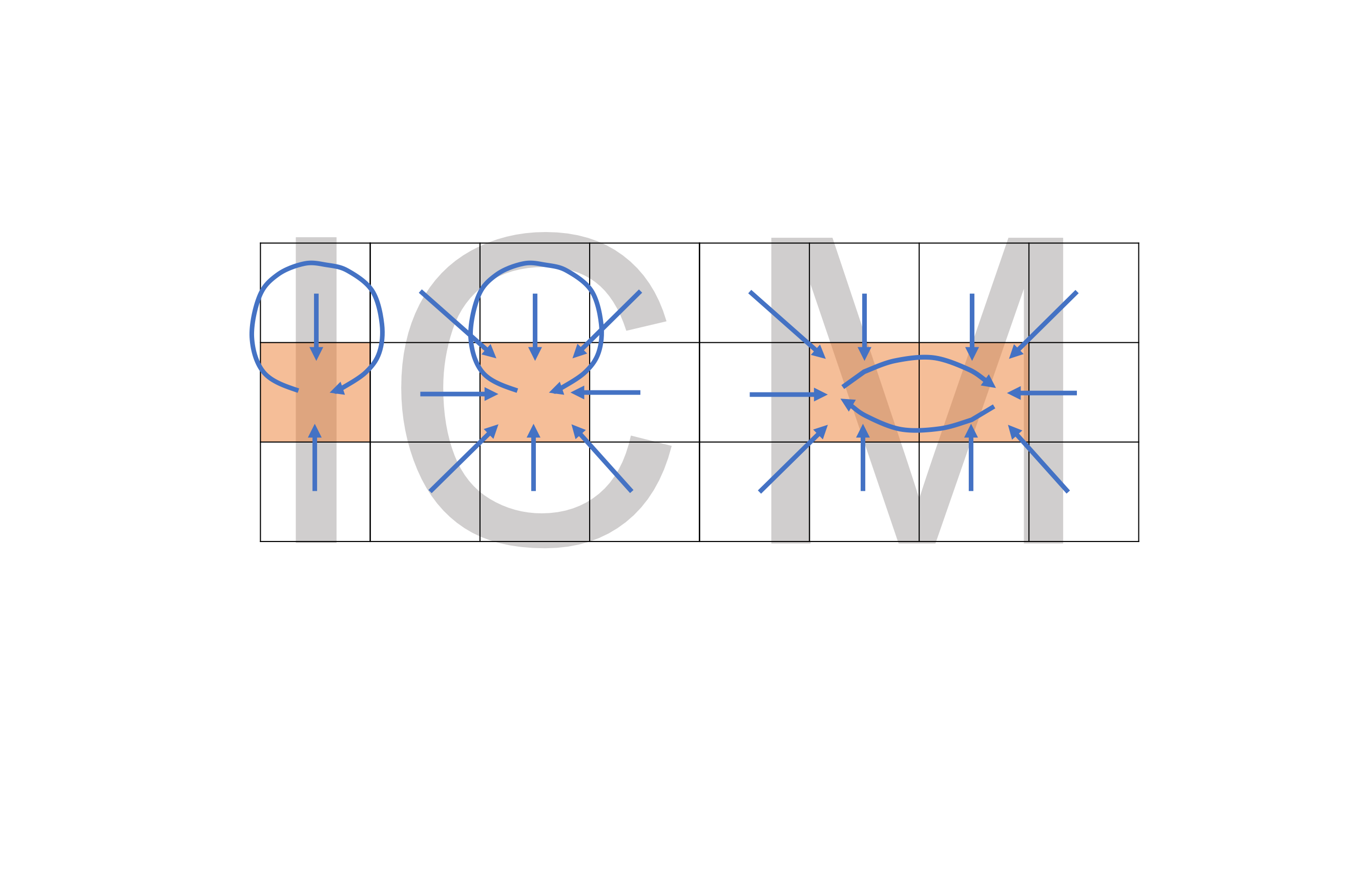}
    \includegraphics[width=0.44\textwidth]{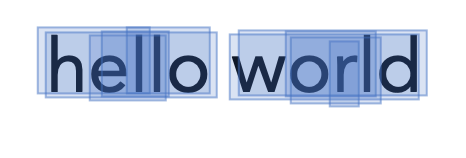}
    \caption{Top: Conceptual visualization of the Graphcore algorithm that enables efficient filtering of character boxes (this is a simplified scenario for representation purposes only). Bottom: constructing word boxes from characters.}
    \label{fig:postpro}
\end{figure}

\subsubsection{Filtering character boxes \emph{efficiently}}
\label{sec:filter-boxes}
Each pixel in the box mask, $B_c$, that  surpasses a certain threshold (e.g. 50\%) produces a candidate character box. The center of the box and its height and width are determined through the model output at the corresponding pixel in ($X_c, Y_c, W_c, H_c$), which is a common procedure in single-stage object detection \cite{SSD, YOLOv3}.
However, this gives multiple boxes around the same character. In order to delete redundantly predicted box proposals of the same character instance, usually NMS is applied~\cite{SSD, FasterRCNN, lin2017focal, he2017mask}. NMS algorithm is quadratic in the number of candidate box proposals. Due to having a large number of box proposals, typically on the order of $10^5$ on a single page, in our case, this becomes computationally expensive. 

To speed up the process, we introduce a preliminary step before NMS, which we call \emph{Graphcore}. Recall, from section~\ref{sec:method}, that each candidate pixel predicts the offset from itself to the center of the predicted character box that lies on that pixel ($X_c, Y_c$). We construct a directed graph where each vertex is a candidate pixel and we add a directed edge going from pixel $A$ to pixel $B$ if pixel $A$ predicts pixel $B$ as the center of its predicted character box. An example graph is visualized in Fig.~\ref{fig:postpro}, top. As can be seen, all the pixels point to the pixels close to the center of the characters and the center pixels point to themselves. In less ideal cases, they may point to a different pixel close to the center.

By taking the k-core of the resulting graph with $k$=1, only the loops in the graph are retained~\cite{DBLP:journals/corr/cs-DS-0310049}. 

This can be performed efficiently in linear-time.
With this, only pixels towards the center of a bounding box, typically one or two candidate boxes per character, are retained. In Fig.~\ref{fig:postpro}, top, these remaining pixels are marked in orange. This significantly smaller set, usually the same order as the number of characters, is passed through NMS to obtain the final character boxes.

\subsubsection{Constructing word boxes from character boxes}
\label{sec:cluster-chars}
Recall from section~\ref{sec:method} that each pixel predicts the center of the word that lies on that pixel. 
Further, from the previous step (Sec.~\ref{sec:filter-boxes}), we have discrete character instances and their boxes. We now wish to cluster the characters into words. Each character makes a word box proposal (for the word it belongs to) based on the word centers predicted by the pixels inside the character; the proposal is such that it extends from the character itself to its reflection on the other side of the predicted word center. 
Fig.~\ref{fig:postpro} shows the word box proposals made by each character in the word \emph{hello world}. The characters closer to the predicted word center propose a smaller word box while the characters farther away from the predicted word center propose a larger word box.

The word proposals, thus generated, significantly overlap for characters belonging to the same word and do not or marginally overlap otherwise. We wish to find clusters of characters whose word proposals significantly overlap. In order to do this, we build a graph where each character is a vertex and the edge between a pair of characters indicates whether their proposals significantly overlap, i.e.~intersection is more than 50\% of the smaller box. We then cluster the characters by finding connected components in this graph. Note that connected component analysis is linear in number of characters. Moreover, this way of clustering characters into words naturally allows us to recognize rotated words.

\section{Building large-scale training datasets}
\label{sec:training-data}
Chargrid-OCR predicts character segmentation mask (i.e. chargrid), character box detection mask, character box centers, widths and heights and word box centers. In order to build the necessary training targets for all the outputs, we need character and word boxes and their string contents. Further, given just the word boxes and their contents, one can closely approximate character boxes and their contents. Therefore, we build our datasets as $(document, words)$ pairs, i.e. for each document, we collect the list of words (boxes and string contents) contained in that document.

To the best of our knowledge, large-scale datasets of documents and word annotations do not exist. While manually collecting word annotations is an option, annotating hundreds of words per document and at least tens of thousands of such documents can be extremely challenging. We, instead, build two imperfect but complementary datasets without resorting to any manual annotations. Below, we describe the two datasets in more detail.

\subsection{Synthetic documents with clean labels}
We generate a dataset by synthetically rendering pdf pages in A4 format using English Wikipedia content. Each page contains parts of a Wikipedia article, with three different font specifications for captions, links and normal text respectively. Each font is sampled randomly from 51 publicly available fonts with varying size and color. For $30\%$ of the pages, we replace $2\%$ of the words with a random string in order to emphasize accurate context-free predictions. Pages are generated with single-column, 2-column or 3-column layout and variable text alignment and figures and tables may be included. Each pdf page can be converted into our model-specific format (i.e. $S$, $B_c$, $X_c$, $Y_c$, $W_c$, $H_c$, $X_w$, $Y_w$), retaining perfect ground truth. The pdf files are converted into grayscale png images with 300dpi.

To more closely mimic a real-world dataset, we perform data augmentation. We consider the following steps (the effects marked with a star* are based on the open source \texttt{ocrodeg} package\footnote{https://github.com/NVlabs/ocrodeg}): 
\emph{(1) Background}: Natural images, gradient background, multiscale noise*, fibrous noise*, blobs*.
\emph{(2) Distortions}: Large 2D distortions*, Small 1D distortions*.
\emph{(3) Projective transformations}: Including rotation, skew, dilation, 3D perspective, etc.
\emph{(4) Degradations}: Gaussian or box blur; mode or median filters; contour, emboss, edges, smooth, gradient text.
\emph{(5) dpi and compression}: Down-scaling, jpeg compression.
\emph{(6) Color}: Equalize, Invert, Sharpness, Contrast, Brightness.
A subset of these steps are randomly chosen and applied on any given document. With this, we generate 66,481 pages of synthetic document data.

\subsection{Real documents with noisy labels}
EDGAR (Electronic Data Gathering, Analysis, and Retrieval) system collects and indexes financial documents submitted by companies to U.S. Securities and Exchange Commission (SEC)\cite{edgar}. 

We create a training dataset by first filtering repeating forms to avoid repeated layouts and text. We filter according to the rule 
$(1000 KB < file\_size < 8000 KB)$ and $(num\_pages > 10)$. This eliminates most duplicated forms and yields sufficient document diversity. From each obtained pdf file, we sample two pages. In total, this yields 42,918 pages.
We obtain ground truth by running Tesseract v4 on each page. 
This results in word bounding boxes for each word as well as the contained string. Character boxes are constructed by splitting the word boxes in proportion to the average width of each character. Since Tesseract v4 produces errors, the labels are highly noisy. To improve the quality, we remove words with very low confidence as well as those with a unlikely width and height. Furthermore, some unicode characters are replaced by ASCII-code equivalents.

\begin{figure*}
    \centering
    \includegraphics[width=0.49\textwidth, height=0.45\textheight, trim={2cm 2cm 2cm 1cm},clip]{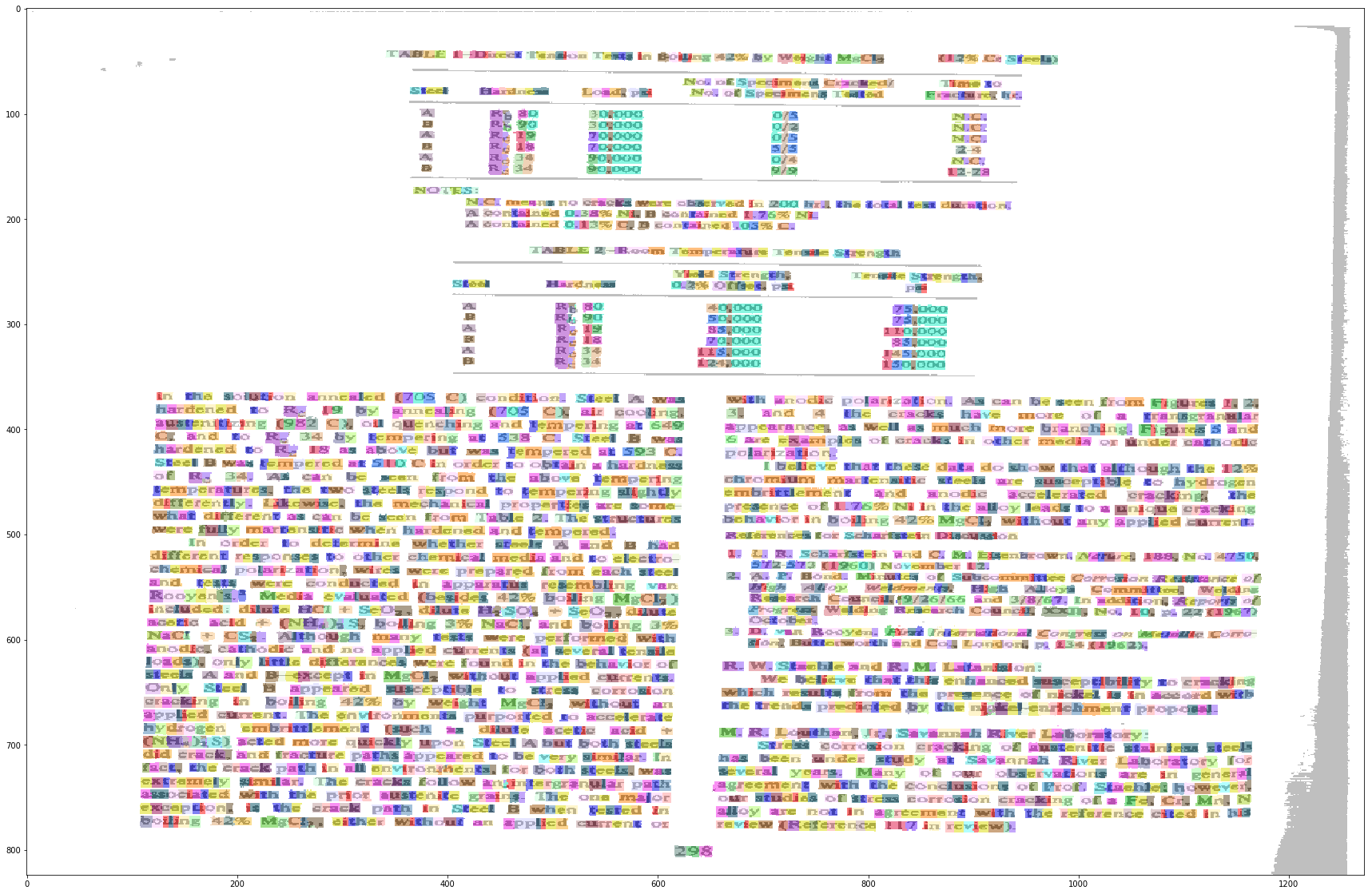}
    \includegraphics[width=0.49\textwidth, height=0.45\textheight, trim={2cm 4cm 2cm 2cm},clip]{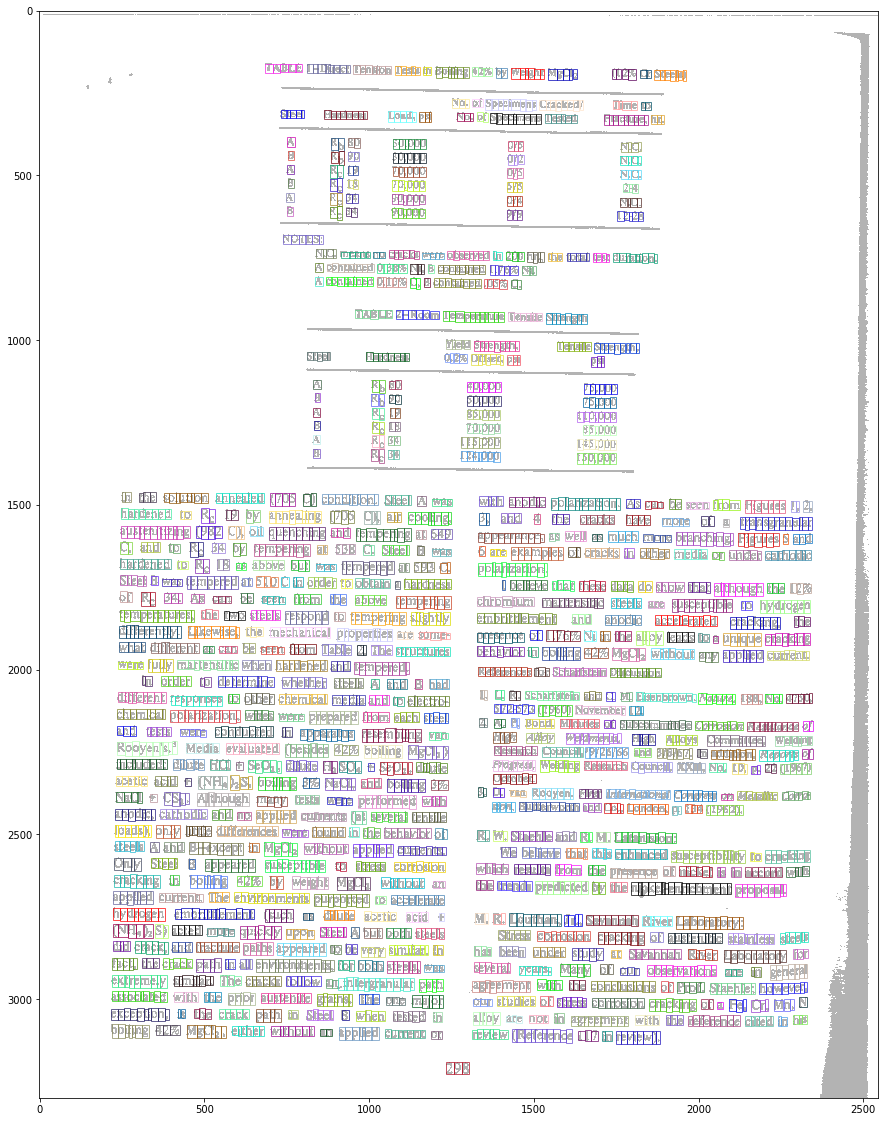}
    \caption{Output of the chargrid-OCR model on a document from DOE Tables dataset~\cite{UNLVtable2010}. Left: The pixel-wise segmentation, i.e. chargrid, with each color representing the predicted character index. Right: The extracted character boxes after postprocessing where the bounding box of each character belonging to the same word is depicted in same color.}
    \label{fig:ex_full_out}
\end{figure*}

\section{Experiments}
In this section, we discuss the evaluation datasets, metric and results. Under results, we first visualize some qualitative examples, followed by comparative results. We then demonstrate the advantages of Chargrid-OCR's end-to-end training by performing domain adaptation and finally analyze the model errors.

\subsection{Evaluation Datasets}
We use two standard document OCR datasets for evaluation, provided by DOE and UNLV. They are both described in \cite{UNLV_bus_letters}.

\textbf{English Business letters.}
This dataset consists of (mostly) free text business letters printed in a variety of fonts. We improved the available ground truth by manually adding word bounding boxes. In total, this dataset consists of 179 pages and 48,687 words.

\textbf{DOE tables.}
We selected the subset of documents containing tables and otherwise structured text. 
We use the ground-truth for the word bounding boxes from \cite{UNLVtable2010}.
In total this dataset consists of 383 pages and 133,245 words.

\subsection{Evaluation Metric}
Document OCR techniques have traditionally been evaluated with layout/line segmentation metrics~\cite{breuel2017robust} 
and text recognition metrics~\cite{breuel2013high, breuel2017high}. 
This is because the solutions for document OCR have been primarily composed of layout/line segmentation and text recognition.
However, in many scenarios, such as in documents with no fluent text (e.g. receipts, invoices, medical reports) and with tabular structures, lines may be ill-defined. Therefore, we choose an evaluation metric that is instead based on words. Further, we wish to capture the location of words in the same metric in addition to the string content. To this end, we use the Word Recognition Rate (WRR) adapted to incorporate the word location.

A predicted word is said to match a ground truth word if and only if their contents agree (i.e. identical strings) and they have a non-zero overlap. Word Recognition Rate can then be computed as $N_m/\big(N_m + N_u + N_g\big)$, where $N_m$ is the number of matched word predictions, $N_u$ is the number of unmatched word predictions and $N_g$ is the number of unmatched (i.e. missed) ground truth words. 
The metric is aggregated across the dataset by taking the average value per document weighted by the number of ground-truth words in the document. This way documents with more text are given higher weight over documents with less text.

\subsection{Results}
\subsubsection{Qualitative results}

Fig.~\ref{fig:ex_full_out} visualizes the outputs of our model on one example from the DOE Tables dataset. The left image visualizes the segmentation mask (chargrid), $S$. Different colors are assigned to different characters for intuitive visualization. As one can observe, character segmentation can easily distinguish textual and non-textual regions. Further, the characters are usually crisp in the simpler regions of the document.

The image on the right visualizes character boxes after \emph{Graphcore} and NMS. Notice the number of character instances on the page and their sizes, particularly thin characters such as \emph{i}, \emph{j}, \emph{t} and \emph{1}. This makes the task of document OCR highly challenging and our task, \emph{ultra-dense} instance segmentation. Clustering characters into words is performed as described in section~\ref{sec:cluster-chars}. The clustering is visualized by showing the characters belonging to the same word in the same color.

\subsubsection{Effect of training datasets}
In Sec.~\ref{sec:training-data}, we highlight the lack of large-scale datasets for document OCR. As a result, we build two imperfect but complementary and large-scale datasets without resorting to any manual annotations. The first dataset comprises of synthetic documents generated from Wikipedia content, the Wiki dataset. This has clean labels. The second dataset consists of real scans of financial reports, the EDGAR dataset, with labels coming from the current most popular OCR solution, Tesseract v4.

In this section, we train our model, Chargird-OCR on the Wiki dataset, the EDGAR dataset and finally, on a combined dataset with examples coming from both Wiki and EDGAR datasets. We choose an input image resolution of 150dpi, i.e. 1648x1272, with the number of convolutional base channels, $C=32$ (see Fig.~\ref{fig:architecture}).
We report the results on the two evaluation datasets in Tab.~\ref{tab:datasets}. It is clear from the results that the two datasets complement each other and that combining them brings significant improvements.

\begin{table}[h]
\centering
  \begin{tabular}{ | l | c | r |}
    \hline
    Training datasets & Letters & DOE tables \\ \hline
    Wiki & 87.5\% & 68.8\% \\ \hline
    EDGAR & 90.4\% & 73.5\% \\ \hline
    Wiki+EDGAR & 90.7\% & 77.6\% \\
    \hline
  \end{tabular}
  \caption{WRR on evaluation sets while training on only one or both training datasets.}
  \label{tab:datasets}
\end{table}

\subsubsection{Comparative results and Timing analysis}

We train various versions of our model on Wiki+EDGAR described in Sec.~\ref{sec:training-data} and report results in Tab.~\ref{tab:results}.
Further, we compare against Tesseract, v3 and v4, with v4~\cite{tesseractgit} (released Oct 2018) being the publicly available state-of-the-art for document OCR. Commercial solutions had to be excluded from evaluation as licenses prevent us from analyzing their results to avoid possible reverse engineering. 

Tesseract v4 comes with an LSTM-based line recognition engine and achieves much higher accuracy than v3. 
Unfortunately, re-training Tesseract on our datasets is not possible due to needing intermediate annotations to train Tesseract. We, therefore, use off-the-shelf Tesseract without any retraining or fine tuning. 
However, we use test data that are domain-independent from training and/or validation data and serve as an indicator for model generalizability.

Our models are dubbed as \emph{ChOCR-$C$}, where $C$, shown in Fig.~\ref{fig:architecture}, determines the width of the network (number of channels in each layer). We also found it important to vary the input image resolution. Therefore, we report at different input DPI.
Tesseract v4 guidelines recommend that it performs best when images are scaled to 300dpi. Therefore, we report on 300dpi resolutions for Tesseract v4 and 150dpi, i.e. (1648 x 1272) and 300dpi, i.e. (3296 x 2544), for Chargrid-OCR; in both cases, the output resolution of Chargrid-OCR is fixed to 150dpi in the $x$ direction and 75dpi in the $y$ direction.

The last column reports time to run 1000 pages through the respective OCR systems. There can be many possible set-ups, thereby making it challenging. However, we choose the simplest set-up - for Tesseract, we run the system on 1 Xeon E5-2698 CPU core and Chargrid-OCR on 1 V100 GPU (plus one CPU core).

It can be seen from Tab.~\ref{tab:results} that \emph{ChOCR-32} at 150dpi is already on-par with or better than Tesseract v4. Due to being able to parallelize on GPUs, our model is $116\times$ faster than Tesseractv4 when executed on a GPU. When we increase the input resolution from 150dpi to 300dpi, the accuracy increases by 2 points. Further increasing the base-channels from 32 to 48 and 64, gives a boost of 2 and 4 points respectively. \emph{ChOCR-64} at 300dpi input resolution is the best performing model; however, different trade-offs between time and accuracy can be chosen.

\begin{table}[t]
\centering
\begin{tabular}{|l|c||c|c|r|} 
\hline
Model & DPI  & Letters &  \shortstack{DOE \\ tables} &  \shortstack{Time \\ (sec/1K pg)} \\
\hline
Tesseract v3 & 300  & 87.7\% &  72.4\% & 5600\\
Tesseract v4 & 300   & 92.6\% &  76.8\% & 14800\\
\hline
ChOCR-32 & 150 & 90.7\% &  77.6\% & 127\\
ChOCR-32 & 300 & 91.7\% &  79.3\% & 284\\
ChOCR-48 & 300 & 92.5\% &  80.4\% & 523\\
ChOCR-64 & 300  & \textbf{93.5\%} &  \textbf{82.0\%} & 795 \\
\hline
\end{tabular}
\caption{Results, reported in terms of Word Recognition Rate. Tesseract run-times are obtained using 1 Xeon E5-2698 CPU core and Chargrid-OCR's on 1 V100 GPU (plus 1 CPU core).}
\label{tab:results}
\end{table}

\begin{figure}
    \centering
    \includegraphics[width=0.23\textwidth, trim={2cm 3cm 1cm 6cm},clip]{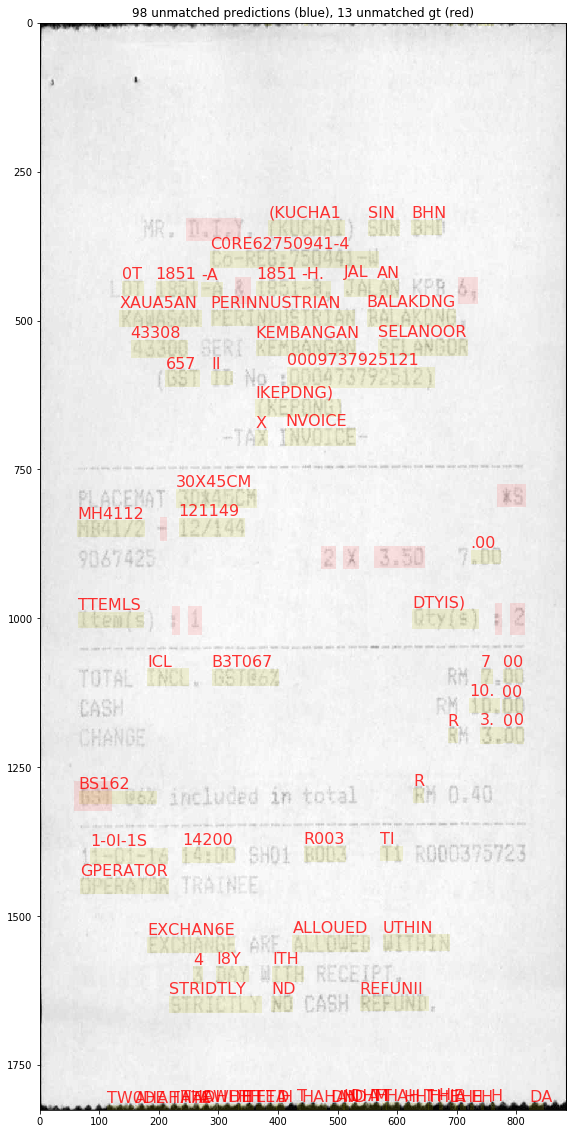}
    \includegraphics[width=0.23\textwidth, trim={2cm 3cm 1cm 6cm},clip]{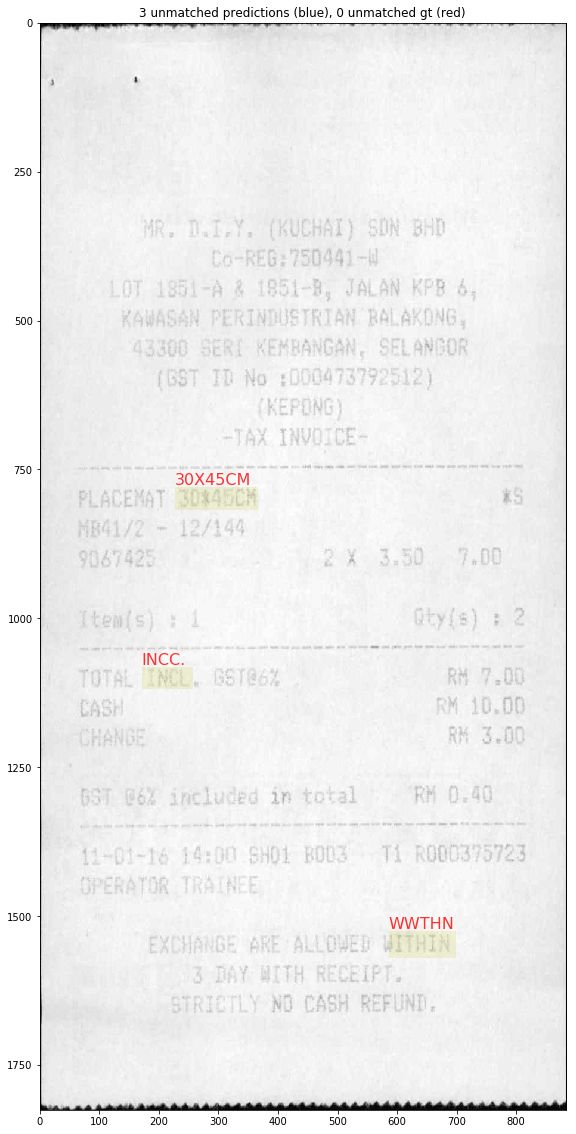}
    \caption{Errors on receipts (SROIE), before and after domain adaptation.}
    \label{fig:receipts}
\end{figure}

\begin{table}[h]
\centering
\begin{tabular}{|l|c|c|c|c||r|} 
\hline
\#examples &  0 &  50 & 100 & 600 (Full) & Tess4 \\
\hline
WRR  & 55.6 & 71.8 & 80.7 & 85.2 & 54.8 \\
\hline
\end{tabular}
\caption{Domain Adaptation on SROIE (Receipts) (\emph{0 examples} means no fine-tuning). Tesseract v4 is shortened to Tess4.}
\label{tab:domain_adaption}
\end{table}

\begin{figure*}
    \centering
    \includegraphics[width=0.49\textwidth, height=0.51\textheight, trim={4cm 4cm 4cm 2cm},clip]{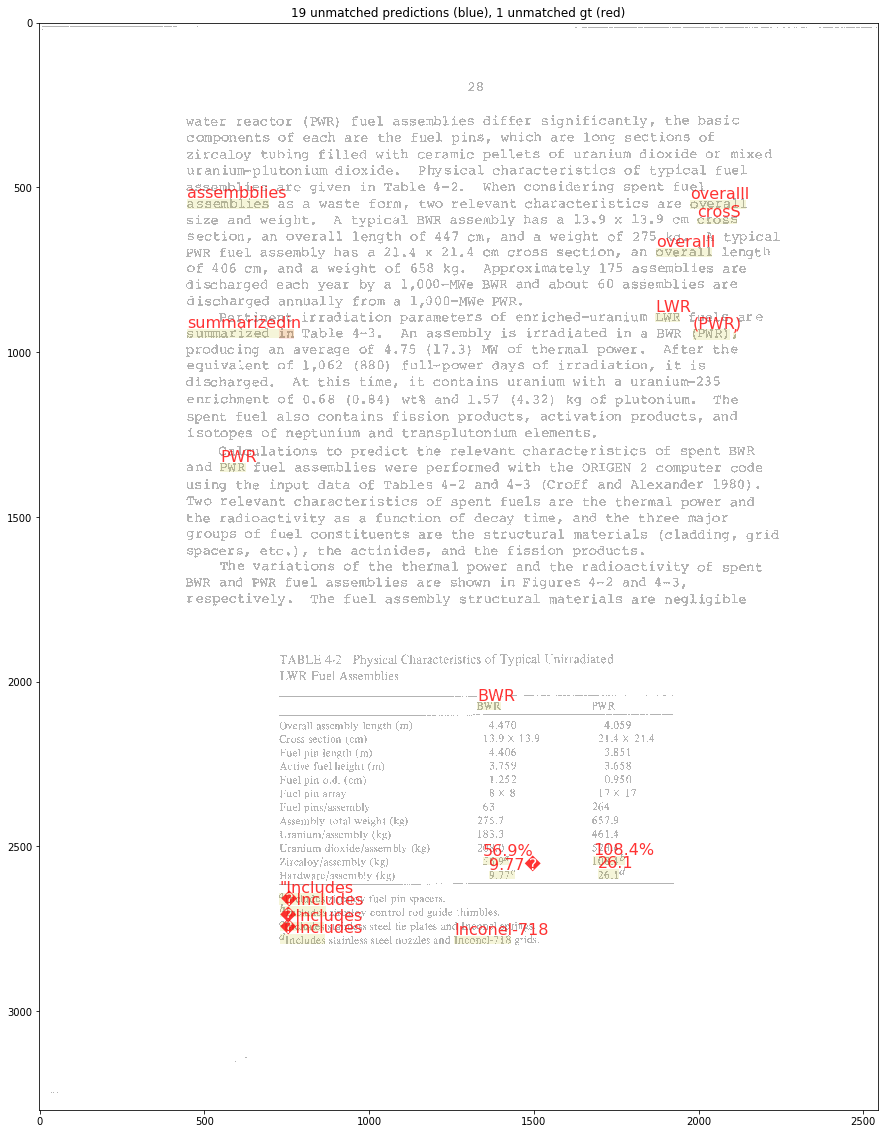}
    \includegraphics[width=0.49\textwidth, height=0.51\textheight, trim={4cm 4cm 4cm 2cm},clip]{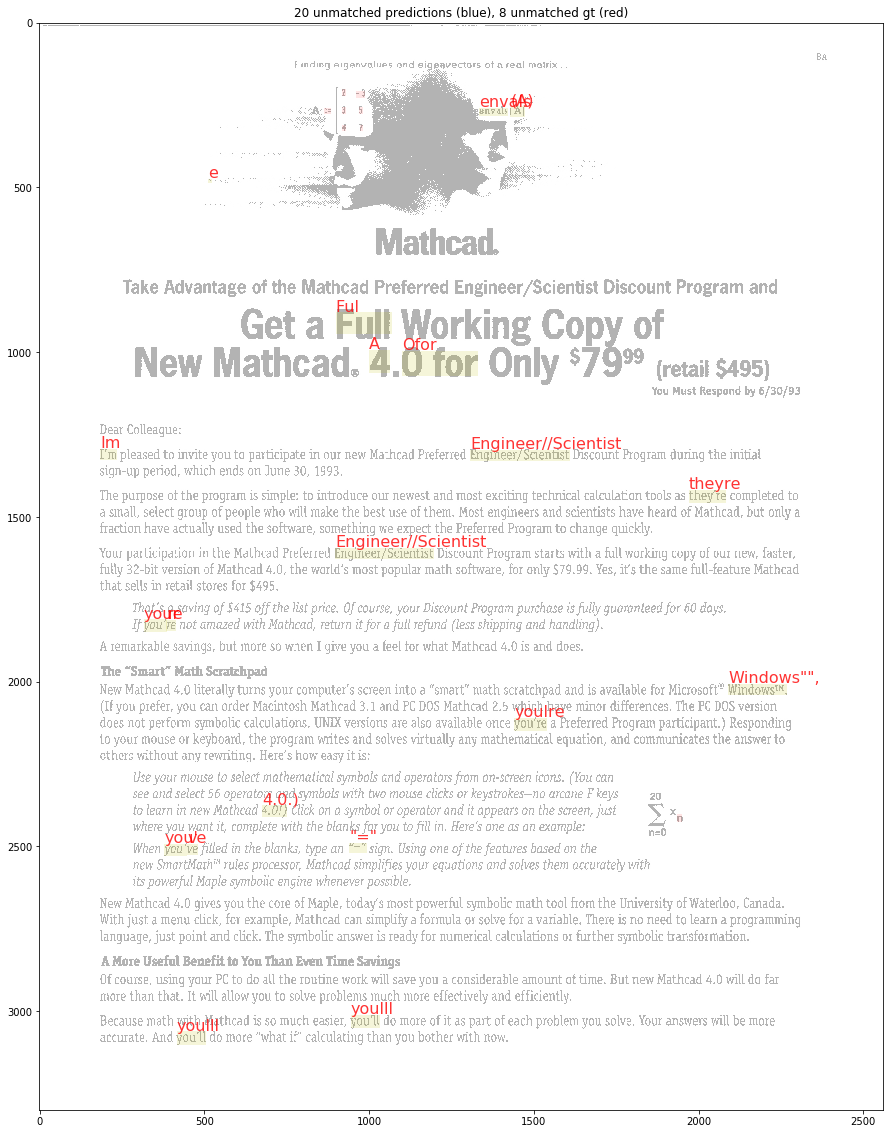}
    \caption{Errors from DOE Tables (left) and Letters (right). The words \emph{LWR, PWR, BWR} in the left image are ground-truth errors.}
    \label{fig:more_ex}
\end{figure*}

\subsubsection{Domain adaptation}
In many scenarios, a model trained on the source domain under-performs on the target domain. Examples of such scenarios are previously unseen document layouts, fonts, noise and domain-specific vocabulary. In such cases, being able to easily adapt the model to the target domain can be very useful. 
In this section, we exploit Chargrid-OCR's end-to-end training capabilities and perform domain adaption via fine-tuning. We use the Scanned Receipts OCR and Information Extraction (SROIE) dataset~\cite{sroie}. This dataset consists of 1000 receipts with OCR annotations, with 600 receipts reserved for training and validation and 400 for test. The domain of receipts is significantly different than printed A4 documents (Fig~\ref{fig:receipts}), with the main differences being font type and the noise.

We evaluate our model, \emph{ChOCR-64}, with no fine-tuning). We then fine-tune our model with 50 , 100 and 600 (full dataset) training samples. 
We evaluate the models and report the results in Tab.~\ref{tab:domain_adaption}. The model without fine-tuning is shown as fine-tuning with \emph{0 examples} for brevity. We also report the results of Tesseract v4 for comparison. 
It can be seen that the accuracies of our model without fine-tuning as well as that of Tesseract v4 considerably drop when compared to the results on A4 documents (Tab.~\ref{tab:results}). This confirms the domain shift. Fine-tuning our model with just 50 example receipts already boosts the WRR by 16 points. Fine-tuning with 100 and 600 examples gives further improvements. 
Fig.~\ref{fig:receipts} visualizes an example receipt and the errors from our model before and after fine-tuning.

\subsubsection{Error analysis}

In Fig.~\ref{fig:more_ex}, we visualize some example A4 documents and the errors of our model. The errors can be categorized as either \emph{character}-related or \emph{word}-related. In case of characters, the following errors are common: (i) as expected, similar looking characters are confused (e.g i-j-t, 1-7-/, etc.); further, we observe a systematic confusion within the sub-categories of numbers, small letters and capital letters indicating some sort of language knowledge learned by the model (ii) thin characters are sometimes missed in a word (e.g. \emph{``helo''}) or repeated more than they appear on the document (referred to as stuttering in the literature); in our model, this can happen due to inaccurate character box detection (iii) in case of very big fonts unseen in the training dataset (such as a page with only the book/chapter title on it), characters can be missed altogether; to make the model more robust against font-size variations, we perform data augmentation with random scaling of documents during training (iv) rare characters (including special characters, subscripts and superscripts) that are not clearly visible on the document are sometimes misclassfied as more frequently occurring characters; this is tackled to an extent with data augmentation using our synthetic dataset, Wiki-dataset.

In case of words, the most common errors come from splitting or merging neighboring words in a different way than the ground-truth. A common example is \emph{``\$  23.45''}, where \emph{``\$''} and \emph{``23.45''} are split into two words whereas in the ground-truth, it may be a single word. Other examples are \emph{``No  :  ''}, \emph{``12. 76''}. However, such errors are often debatable even by humans. Further, there are ground-truth inconsistencies leading to inconsistent predictions.

Within Chargrid-OCR, we do no include a language model. This leads to predictions that are more \emph{``as seen''} than \emph{``as read''}. This was rather a design choice to exclude the language model from the recognition model. However, this also leads to errors that a human may consider easy. Further, while such errors appear more prominently in fluent text such as Letters dataset, the lack of a language model produces unbiased output in documents with no fluent text and with tabular structures such as the DOE tables dataset.

\section{Conclusion}
\label{sec:conclusion}

We presented a new end-to-end trainable optical character recognition system for printed documents that is based on character instance segmentation.
Our model is trained on a combination of synthetically generated documents with clean labels and real documents with noisy labels coming from the current state-of-the-art document OCR solution.
We empirically show that on Business Letters and DOE Tables datasets for OCR, we outperform the current state-of-the-art both in terms of accuracy and run-time while being significantly easier to train and adapt.

{\small
\bibliographystyle{ieee_fullname}
\bibliography{egbib}
}

\end{document}